\documentclass{article}
\usepackage{jmlr2e}
\usepackage{amsmath,amssymb}

\usepackage{graphicx}

\newcommand{\mysec}[1]{Section~\ref{sec:#1}}
\newcommand{\eq}[1]{Eq.~(\ref{eq:#1})}
\newcommand{\myfig}[1]{Figure~\ref{fig:#1}}

\newcommand{\BEAS}{\begin{eqnarray*}}
\newcommand{\EEAS}{\end{eqnarray*}}
\newcommand{\BEA}{\begin{eqnarray}}
\newcommand{\EEA}{\end{eqnarray}}
\newcommand{\BEQ}{\begin{equation}}
\newcommand{\EEQ}{\end{equation}}
\newcommand{\BIT}{\begin{itemize}}
\newcommand{\EIT}{\end{itemize}}
\newcommand{\BNUM}{\begin{enumerate}}
\newcommand{\ENUM}{\end{enumerate}}
\newcommand{\BA}{\begin{array}}
\newcommand{\EA}{\end{array}}

\newcommand{\cov}{\mathop{\rm cov}}

\newcommand{\rb}{\mathbb{R}}

\newcommand{\idm}{I}

\title{Graph kernels between point clouds}

\author{\name Francis R. Bach \email francis.bach@mines.org\\
\addr
INRIA - Willow project\\
D\'epartement d'Informatique, Ecole Normale Sup\'erieure\\
45, rue d'Ulm\\
75230 Paris, France\\
       }

\begin{document}

\maketitle

\begin{abstract}
Point clouds are sets of points in two or three dimensions. Most kernel methods for learning on sets of points have not yet dealt with the specific geometrical invariances and practical constraints associated with point clouds in computer vision and graphics. In this paper, we present extensions of graph kernels for point clouds, which allow to use kernel methods for such objects as shapes, line drawings, or any three-dimensional point clouds.
In order to design rich and numerically efficient kernels with as few free parameters as possible, we use kernels between covariance matrices and their factorizations on graphical models. We derive polynomial time dynamic programming recursions and  present applications to recognition of handwritten digits and Chinese characters from few training examples.

\end{abstract}

\section{Introduction}

In recent years, kernels for structured data have been designed in many domains, such as bioinformatics~\cite{vert}, speech processing~\cite{vert_speech}, text processing~\cite{lodhi} and computer vision~\cite{harchaoui}. They provide an elegant way of including known \emph{a priori} information, by using directly the natural topological structure of objects. Using a priori knowledge through structured kernels have proved beneficial because it allows to reduce the number of training examples, and to re-use existing data representations that are already well developed by experts of those domains.

In this paper, we propose a kernel between point clouds, with applications to classification of line drawings (such as handwritten digits~\cite{lecun-98} or Chinese characters~\cite{chinese,Sun_05a}) or shapes~\cite{shape}. 
The natural geometrical structure of point clouds is hard to represent in a few real-valued features~\cite{ponce}, in particular because of (a)  the required local or global invariances by rotation, scaling, and/or translation, (b) the lack of pre-established registrations of the point clouds (i.e., points from one cloud are not matched to points from another cloud), and (c) the noise and occlusion that impose that only portions of two point clouds ought to be compared.

Following one of the leading principles for designing kernels between structured data, we propose to look at all possible partial matches between two point clouds~\cite{ShaweCristi04}. More precisely, we assume that each point cloud has a graph structure (most often a neighborhood graph), and we consider recently introduced \emph{graph kernels}~\cite{RamonGaert03,KashiTsudaInoku04}. Intuitively, these kernels consider all possible subgraphs and compare and count the matching subgraphs. However, the set of subgraphs (or even the set of paths) has exponential size and cannot be efficiently described recursively; so larger sets of substructures are commonly used, e.g., walks and tree-walks.  As shown in \mysec{graphkernels}, by choosing appropriate substructures and fully factorized local kernels, efficient dynamic programming implementations allow to sum over an exponential number of substructures in polynomial time. The kernel thus provides an efficient and elegant way of considering very large feature spaces (see, e.g.,~\cite{ShaweCristi04}).

However, in the context of computer vision, substructures correspond to matched sets of points, and dealing with local invariances imposes to use a local kernel that cannot be readily expressed as a product of separate terms for each pair of points, and the usual dynamic programming approaches cannot then be applied.
The main contribution of this paper is to design a local kernel that is not fully factorized but can be instead factorized according to the graph underlying the substructure. This is naturally done through graphical models and the design of positive kernels for covariance matrices that factorize on graphical models (\mysec{GM}). With this novel local kernel, we derive new polynomial time dynamic programming recursions in \mysec{recursions}. In \mysec{simulations}, we present simulations on handwritten character recognition.

\section{Graph kernels}
\label{sec:graphkernels}
In this section, we consider two labelled undirected graphs $G=(V,E,a,x)$ and $H=(W,F,b,y)$. Two types of labels are considered: \emph{attributes}, which are denoted $a(v) \in \mathcal{A}$ for vertex $v\in V $  and $b(w)\in \mathcal{A}$ for vertex $w \in W$ and \emph{positions}, which are denoted $x(v) \in \mathcal{X}$ and $y(w) \in \mathcal{X}$. Our motivating examples are line drawings, where $\mathcal{X} = \mathcal{A} = \mathbb{R}^2$.

The definition of  graph kernels between $G$ and $H$ relies on a set of substructures of the graphs. The most natural ones are paths, subtrees and more generally subgraphs; however, they do not lead to efficient enumerations, and recent work~\cite{RamonGaert03,mahe-tech} has focused on larger sets of substructures that we now present.

\subsection{Paths, walks, subtrees and tree-walks}
\begin{figure}
\begin{center}
\includegraphics[scale=.32]{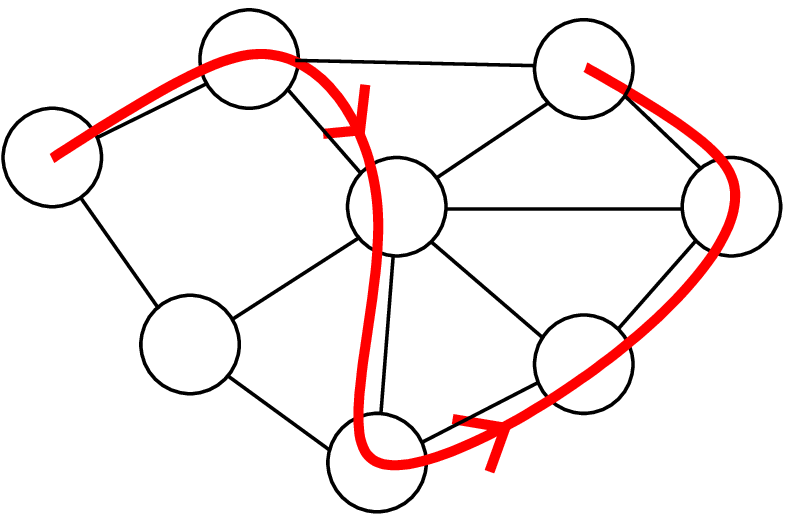} \hspace*{1cm}
\includegraphics[scale=.32]{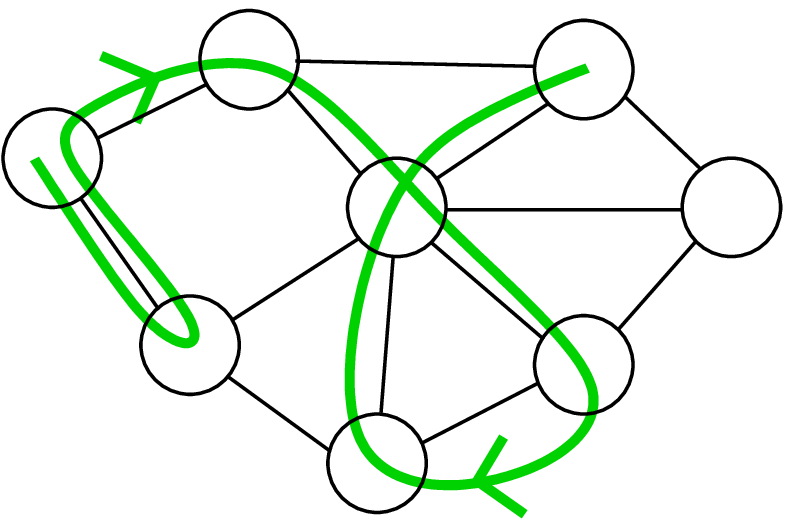} \hspace*{1cm}
\includegraphics[scale=.32]{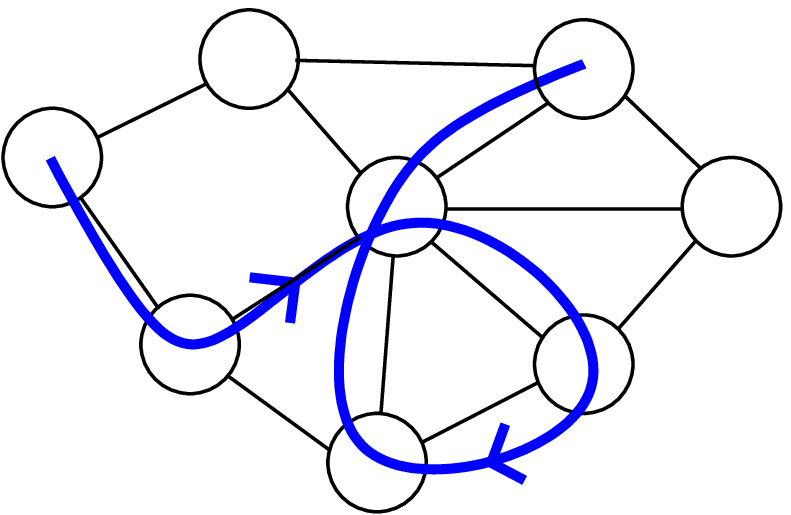} 

\vspace*{-.25cm}

\caption{(left) path, (center) $1$-walk which is not a $2$-walk, (right) $2$-walk which is not a $3$-walk.
 }
 \label{fig:walks}
\end{center}
\end{figure}

Given the undirected graph $G$ with vertex set $V$, a \emph{path} is a sequence of distinct connected vertices, while a \emph{walk} is a sequence of possibly non distinct connected vertices. For any positive integer $\beta$, we define $\beta$-walks as walks such that any $\beta+1$ successive vertices are distinct ($1$-walks are regular walks). Note that when the graph $G$ is a tree (no cycles), then the set of $2$-walks is equal to the set of paths (see examples in \myfig{walks}). More generally, for any graph, $\beta$-walks of length $\beta+1$ are exactly paths of length $\beta+1$.

A subtree is a subgraph of with no cycles. We can represent a subtree of $G$ by a tree structure $T$ over the vertex set $\{1,\dots,|T|\}$, where $|T|$ is the number of nodes in $T$, and a sequence of  \emph{distinct consistent} labels $I \in V^{|T|}$ (i.e., that are neighbors in $G$ when neighbors in $T$).  In this paper, we consider only \emph{rooted} subtrees, i.e.,  subtrees where a specific node is identified as the root.\footnote{Moreover, all the trees that we consider are unordered trees (i.e., no order is considered among siblings).}

\begin{figure}
\begin{center}
\includegraphics[scale=.32]{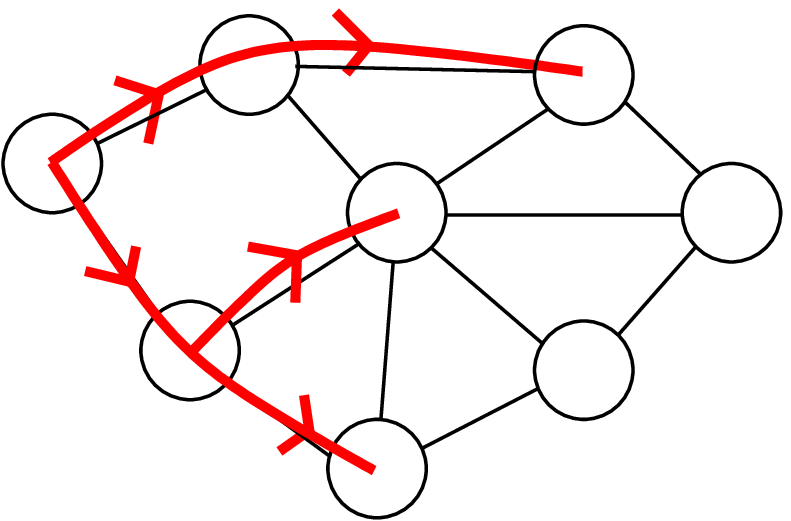} \hspace*{1cm}
\includegraphics[scale=.32]{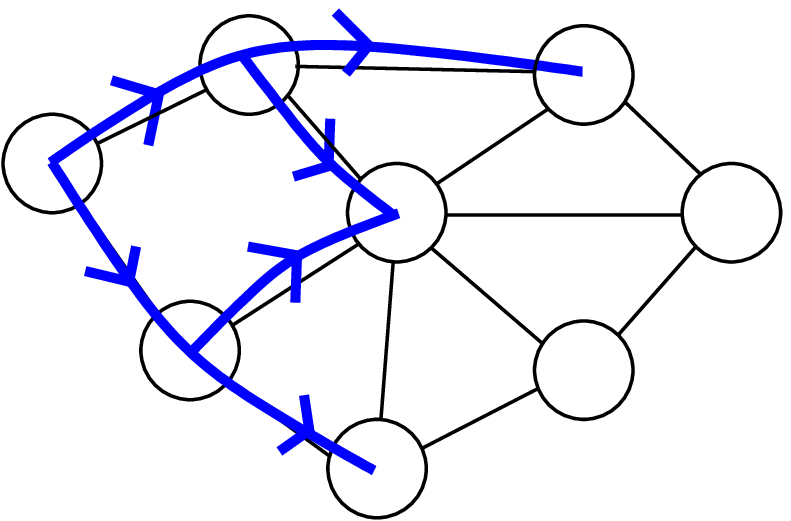} 

\vspace*{-.25cm}

\caption{(left) $2$-tree-walk, (center) $1$-tree-walk which is not a $2$-tree-walk.
 }
 \label{fig:treewalks}
\end{center}
\end{figure}

The notion of walk is extending the notion of path by allowing nodes to be equal. Similarly, we can extend the notion of subtrees to \emph{tree-walks}, which can have nodes that are equal. More precisely, we define an $\alpha$-ary tree-walk of depth $\gamma$ of $G$ as a (non complete) labelled $\alpha$-ary tree of depth $\gamma$ with nodes labelled by vertices in $G$, and such that the labels of neighbors in the tree-walk are neighbors in $G$. A tree-walk may be represented by a tree structure $T$ over the vertex set $\{1,\dots,|T|\}$  and a sequence of consistent but possibly non distinct labels $I \in V^{|T|}$. We can also define $\beta$-tree-walks, as tree-walks such that for each node in $T$, its label (i.e. an element of $V$) and the ones of all its descendants up to the $\beta$-th generation are all distinct. 
With that definition, $1$-tree-walks are regular tree-walks 
(see \myfig{treewalks}).
Note that if $\alpha=1$, we get back $\beta$-walks and the graph kernels that we use are often referred to as \emph{random walk kernels}~\cite{RamonGaert03}. From now on, we refer to the  descendants up to the $\beta$-th generation as the $\beta$-descendants.

We let denote $\mathcal{T}_{\alpha,\gamma}$ the set of tree structures of depth less than $\gamma$ and with at most $\alpha$ children per node. For $T \in \mathcal{T}_{\alpha,\gamma}$, we define $\mathcal{J}_{\beta}(T,G)$ the set of consistent labellings of $T$ by vertices in $V$ leading to $\beta$-tree-walks. With these definitions, a $\beta$-tree-walk of $G$ is characterized by a tree structure $T \in \mathcal{T}_{\alpha,\gamma}$ and a labelling $I \in \mathcal{J}_{\beta}(T,G)$.

\subsection{Graph kernels}
Assuming a local kernel $q_{T,I,J}(G,H)$ between two tree-walks that share the same structure is given, following~\cite{RamonGaert03}, we can define the \emph{tree-kernel} as the sum over all matching tree-walks of $G$ and $H$ of the local kernel:
\BEQ
\label{eq:def}
k^{\mathcal{T}}_{\alpha,\beta,\gamma}(G,H) = \sum_{T \in \mathcal{T}_{\alpha,\gamma}}   f_{\lambda,\nu}(T)
\sum_{I \in \mathcal{J}_{\beta}(T,G) }
\sum_{J \in \mathcal{J}_{\beta}(T,H) }
 q_{T,I,J}(G,H). \EEQ
 If the kernel $q_{T,I,J}(G,H)$ has positive values and is equal to 1 if the two tree-walks are equal, it can be seen as a soft matching indicator, and then the kernel in \eq{def} simply counts the softly matched tree-walks in the two graphs.

We add a nonnegative penalization $f_{\lambda,\nu}(T) $ depending only on the tree-structure. Besides the usual penalization of the number of nodes ${|T|}$, we also add a penalization of the number of leaf nodes $\ell(T)$. More precisely, we use the penalization  $f_{\lambda, \nu} = \lambda^{|T|} \nu^{\ell(T)}$. This penalization suggested by~\cite{mahe-tech} is essential in our situation to avoid that trees with nodes of higher degrees dominate the sum.

If $q_{T,I,J}(G,H)$ is a positive kernel between $G$ and $H$, then $k^{\mathcal{T}}_{\alpha,\beta,\gamma}(G,H)$ also defines a positive kernel. The kernel  $k^{\mathcal{T}}_{\alpha,\beta,\gamma}(G,H)$ sums the \emph{local kernel} $q_{T,I,J}(G,H)$ overall all tree-walks of $G$ and $H$ that share the same tree structure. The number of matching tree-walks is exponential in the depth $\gamma$, thus, in order to deal with potentially deep trees, a recursive definition is needed. It requires a specific type of local kernels.

\subsection{Local kernel}
We use a combination (product) of a kernel for attributes and a kernel for positions. For attributes, we use the following usual
factorized form $ k_\mathcal{A}(a(I),b(J)) = \prod_{p=1}^{|I|} k_\mathcal{A}(a(I_p),b(J_p)$, where $k_\mathcal{A}$ is a positive definite kernel on $\mathcal{A} \times \mathcal{A}$. 
This allows the separate comparison of each matched pair of points and efficient dynamic programming recursions~\cite{RamonGaert03,harchaoui}. However, for our local kernel on positions, we need a kernel that \emph{jointly} depends on the whole vectors $x(I)$ and $y(J)$, and not only on pairs $(x(I_p),y(J_p))$.

In this paper, we focus on $\mathcal{X}=\rb^d$ and translation invariant local kernels, which implies that the local kernel for positions may only depend on differences $x(i)-x(i')$ and $y(j)-y(j')$ for $(i,i') \in I\times I $ and $(j,j')\in J \times J$. We further reduce these to the kernel matrices corresponding to a translation invariant kernel $k_\mathcal{X}(x-x')$.  Depending on the application, $k_\mathcal{X}$ may or may not be rotation invariant. 

That is, we define full kernel matrices $K \in \rb^{|V|\times |V|}$ and $L \in \rb^{|W|\times |W|}$ for each graph, defined as $K(v,v')=
k_\mathcal{X}(x(v)-x(v'))$ (and  similarly for $L$). For simplicity, we assume that these matrices are positive definite (i.e., invertible), which can be enforced by adding a multiple
of the identity matrix. The local kernel will thus only depend on
the submatrices $K_I=K_{I,I}$ and $L_J=L_{J,J}$, which are positive definite matrices. Note that we use kernel matrices $K$ and $L$ to represent the geometry of each graph, and that we use a kernel on such kernel matrices.

We use the following kernel on positive matrices $K$ and $L$, the  (squared) Bhattacharyya kernel $k_{\mathcal{B}}$, defined as~\cite{kondor}:
\BEQ
 \label{eq:bata}
 \textstyle k_{\mathcal{B}}(K,L) = |K|^{1/2} |L|^{1/2} \left| \frac{K+L}{2} \right|^{-1}, 
\EEQ
which is a positive kernel with pointwise positive values and such that $ k_{\mathcal{B}}(K,K)=1$ ($|K|$ denotes the determinant of $K$).

By taking the product of the attribute-based local kernel and the position-based local kernel, we get the following local kernel
$ q_{T,I,J}^0(G,H) = k_{\mathcal{B}}(K_I,L_J)  k_\mathcal{A}(A_I,B_J)$. However, this local kernel $ q_{T,I,J}^0(G,H)$ does not yet depend on the tree structure $T$ and the recursion may be efficient only if $q_T$ can be computed recursively. The factorized term $k_\mathcal{A}(A_I,B_J)$ does not cause any problems; however, for the term $k_{\mathcal{B}}(K_I,L_J)$, we need an approximation based on $T$. As we show in \mysec{GM}, this can be  obtained by a  factorization according to the appropriate graphical model.

\section{Positive matrices and graphical models}
\label{sec:GM}
The main idea underlying the factorization of the kernel is to consider symmetric positive matrices as covariance matrices and look at graphical models defined for Gaussian random vectors with those covariance matrices. In this section we assume that we have $n$ random variables $Z_1,\dots,Z_n$ with probability distribution $p(z)=p(z_1,\dots,z_n)$. Given a kernel matrix $K$ (in our case defined as $K_{ij}=e^{-\alpha\|x_i-x_j\|^2}$, for positions $x_1,\dots,x_n$), we consider  random variables $Z_1,\dots,Z_n$ such that $\cov(Z_i,Z_j) = K_{ij}$. In this section, with this identification, we consider covariance matrices as kernel matrices, and vice-versa.

\subsection{Graphical models and junction trees}
Graphical models provide a flexible and intuitive way of defining factorized probability distributions.  Given any undirected graph $Q$ with vertices in $\{1,\dots,n\}$, the distribution $p(z)$ is said to factorize in $Q$ if it can be written as a product of potentials over all cliques (completely connected subgraphs) of the graph $Q$. When the distribution is Gaussian with covariance matrix  $K \in \rb^{n \times n}$, the distribution factorizes if and only if $(K^{-1})_{ij}=0$ for each $(i,j)$ which is not an edge in $Q$~\cite{lauritzen,jordanreview}.

\begin{figure}
\begin{center}
\hspace*{-.2cm}
\includegraphics[scale=.42]{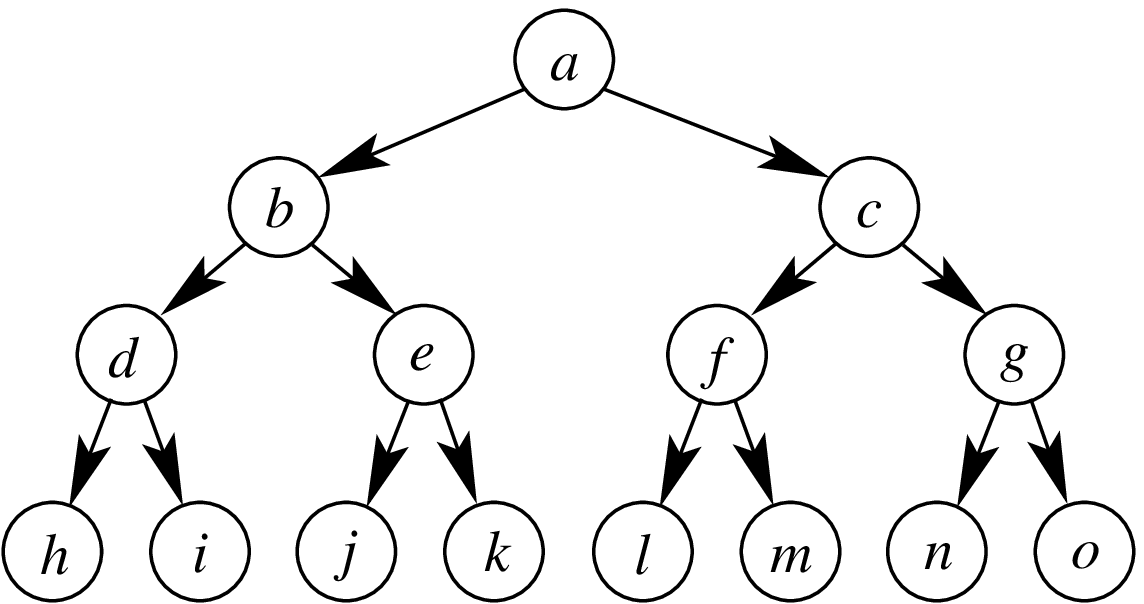}
\hspace*{.25cm}
\includegraphics[scale=.42]{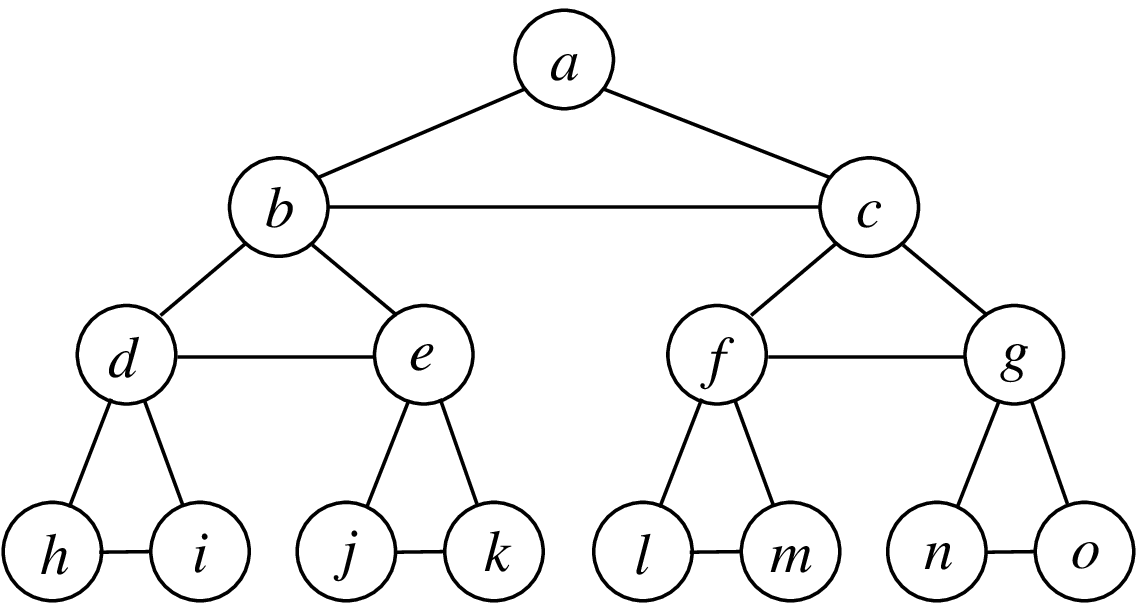}
\hspace*{.25cm}
\includegraphics[scale=.45]{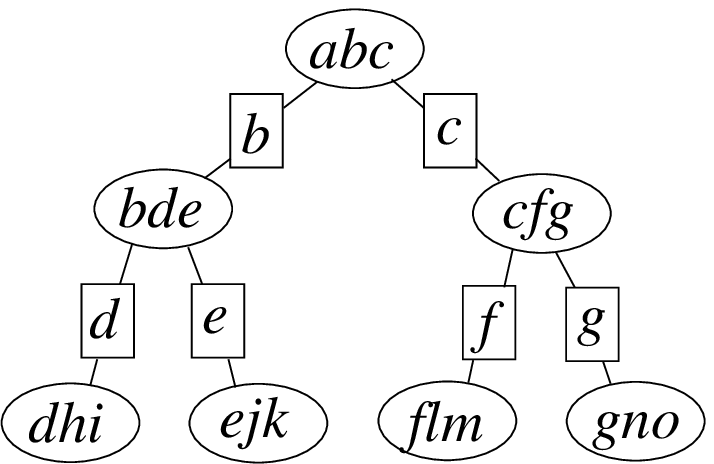}
\hspace*{-.2cm}
\caption{(left) rooted tree structure $T$, (middle) decomposable graphical model $Q_1(T)$, defined in \mysec{GM2}; (right) corresponding rooted junction tree with cliques (ellipses) and separators (rectangles). By convention, all trees are rooted from top to bottom.
 }
 \label{fig:cliques}
\end{center}
\end{figure}

In this paper, we only consider \emph{decomposable} graphical models, for which the graph $Q$ is \emph{triangulated} (i.e., there exists no chordless cycles of length larger than 4). In this case, the joint distribution is uniquely defined from its marginals $p(z_C)$ on the cliques $C$  of the graph $Q$.
Namely, if $\mathcal{C}(Q)$ is the set of maximal cliques of $Q$, we can build a tree of cliques, a \emph{junction tree}, such that $p(z) = \prod_{C \in \mathcal{C}(Q)} p(z_C) / \prod_{C,C' \in \mathcal{C}(Q), C \sim C'}p(z_{C \cap C'})$.
(see \myfig{cliques} for an example of graphical model and junction tree). The sets $C \cap C'$ are usually referred to as \emph{separators} and we let denote $\mathcal{S}(Q)$ the set of such separators.

\subsection{Graphical models and projections}

We let denote $\Pi_Q(K)$ the covariance matrix
that factorizes in $Q$ which is closest to $K$ for the Kullback-Leibler divergence between normal distributions with covariance matrices $K$ and $L$. In this paper, we essentially replace $K$ by $\Pi_Q(K)$; i.e., we project all our covariance matrices onto a graphical model, which is a classical tool in probabilistic modelling~\cite{jordanreview}. We leave the study of the approximation properties of such a projection (in particular along the lines of~\cite{caelli}) to future work.

Practically, since our kernel on kernel matrices involves determinant, we simply need to compute $|\Pi_Q(K)|$ efficiently.
For decomposable graphical models, $\Pi_Q(K)$ can be obtained in closed form~\cite{lauritzen} and its determinant has the following simple expression:
%
\BEQ
\textstyle
\label{eq:det1}
 \log | \Pi_Q(K) | = \sum_{C \in \mathcal{C}(Q)} \log | K_{C}|  - \sum_{S \in \mathcal{S}(Q)} \log | K_{S}|.
\EEQ
If the junction tree is rooted (by choosing any clique as the root), then for each clique but the root, a unique parent clique $p_Q(C)$ is defined, and we have:
\BEQ
\textstyle
\label{eq:det2}
 \log | \Pi_Q(K) |
= 
\sum_{ C \in \mathcal{C}(Q) }  \log  \frac{| K_{C} |  }{| K_{p_Q(C)} | } = \sum_{ C \in \mathcal{C}(Q) }  \log | K_{C | p_Q(C)} |,
\EEQ
where $p_Q(C)$ is the parent clique of $Q$ (and $\varnothing$ for the root clique) and the conditional covariance is defined, as usual, as
$K_{C | p_Q(C)} = K_{C,C}-K_{C,p_Q(C)}K_{p_Q(C),p_Q(C)}^{-1} K_{p_Q(C),C}$.

\subsection{Graphical models and kernels}
We now propose several ways of defining a kernel adapted to graphical models. All of them are based on replacing determinants $|M|$ by  $|\Pi_Q(M)|$, and their different decompositions in \eq{det1} and \eq{det2}.
Using \eq{det1}, we obtain the first kernel:
\BEQ
\label{eq:kernelgraphcov}
\textstyle
 k_{\mathcal{B},0}^Q(K,L) =  \prod_{C \in \mathcal{C}(Q)}  k_\mathcal{B}(K_C,L_C)
 \left(   \prod_{S \in \mathcal{S}(Q)}  k_\mathcal{B}(K_S,L_S) \right)^{-1}.
\EEQ
However, this is not always a positive kernel for general covariance matrices:
\begin{proposition}
For any decomposable model $Q$, the kernel $k_{\mathcal{B},0}^Q$ defined in \eq{kernelgraphcov} is a positive kernel on the set of covariance matrices $K$ such that for all separators $S \in \mathcal{S}(Q)$, $K_{S,S}=\idm$. In particular, when all separators have cardinal one, this is a kernel on correlation matrices.
\end{proposition}

In order to remove the condition on separators, we consider the rooted junction tree representation in \eq{det2} and define another kernel as
\BEQ
\textstyle
\label{eq:q2a}
k_{\mathcal{B}}^Q(K,L) =  \prod_{C \in \mathcal{C}(Q)} k_{\mathcal{B}}^{C|p_Q(C)}(K,L).
\EEQ 
For the root, we define $k_{\mathcal{B}}^{R|\varnothing}(K,L)
=k_\mathcal{B}(K_R,L_R) $ and the kernels $k_{\mathcal{B}}^{C|p_Q(C)}(K,L)$ are defined as kernels between conditional Gaussian distributions of $Z_C$ given $Z_{p_Q(C)}$. We use
\BEQ
\label{eq:q2b}
k_{\mathcal{B}}^{C|p_Q(C)}(K,L)\! = \! \frac{|K_{C|p_Q(C)}|^{1/2} |L_{C|p_Q(C)}|^{1/2}}{  \left| \frac{1}{2} K_{C|p_Q(C)}\!  +  \! \frac{1}{2} L_{C|p_Q(C)}\!  + \!  \frac{1}{4} ( K_{C,p_Q(C)} K_{p_Q(C)}^{-1}\!  - \!  L_{C,p_Q(C)} L_{p_Q(C)}^{-1} )^{\otimes 2} \right| },
\EEQ
which corresponds to putting a prior with identity covariance matrix on variables $Z_{\Pi(Q)}$ and considering the kernel between the resulting joint covariance matrices on  variables indexed by $(C,p_Q(C))$ (we use the notation $M^{\otimes 2} = MM^\top$). We now have a positive kernel on all covariance matrices:

\begin{proposition}
For any decomposable model $Q$, the kernel $k_{\mathcal{B}}^Q(K,L)$ defined in \eq{q2a} and \eq{q2b} is a positive kernel on the set of covariance matrices.
\end{proposition}

Note that the kernel is not invariant by the choice of the particular root of the junction tree. However, in our setting, this is not an issue because we have a natural way of rooting the junction trees (i.e, following the rooted tree-walk). 

In \mysec{recursions}, we will use the notation $k_\mathcal{B}^{I_1|I_2, J_1|J_2}(K,L)$ for $|I_1|=|I_2|$ and $|J_1|=|J_2|$ to denote the kernel between covariance matrices $K_{I_1 \cup I_2}$ and $L_{I_1 \cup I_2}$ adapted to the conditional distributions $I_1|I_2$ and $J_1|J_2$ (defined through \eq{q2b}).

\subsection{Choice of graphical models}
\label{sec:GM2}
Given the rooted tree structure $T$ of a $\beta$-tree-walk, we now need to define the graphical model $Q_\beta(T)$ that we use to project
our kernel matrices.  We define $Q_\beta(T)$ such that for all nodes in $T$, the node together with all its $\beta$-descendants form a clique, i.e., a node is connected to its $\beta$-descendants and all $\beta$-descendants are also mutually connected (see \myfig{cliques} for example for $\beta=1$): the set of cliques are thus the set of \emph{families} of depth $\beta$ (i.e., with $\beta+1$ generations).
 Thus our final kernel is:
\BEQ
\label{eq:kernel}
k^\mathcal{T}_{\alpha,\beta,\gamma}(G,H) = \sum_{T_\in \mathcal{T}_{\alpha,\gamma}} f_{\lambda,\nu}(T)
\sum_{I \in \mathcal{J}_{\beta}(T,G) }
\sum_{J \in \mathcal{J}_{\beta}(T,H) }
 k_{\mathcal{B}}^{Q_\beta(T)}(K_I,L_J) k_\mathcal{A}(A_I,B_J).
 \EEQ
 The main intuition behind this definition is to sum local similarities over all matching subgraphs. In order to obtain a tractable formulation, we simply needed to (a)  extend the set of subgraphs (to tree-walks of depth $\gamma$) and (b) factorize the local similarities along the graphs. Note that the graph $Q_\beta(T)$ that we chose is the densest graph for which the following dynamic programming recursions may hold.

\section{Dynamic programming recursions}

\label{sec:recursions}
In order to derive dynamic programming recursions, we follow~\cite{mahe-tech} and rely on the fact that $\alpha$-ary $\beta$-tree-walks of $G$ can essentially be defined through $1$-tree-walks on the augmented graph of all subtrees of $G$ of depth at most $\beta-1$ and arity less than $\alpha$.

We thus consider the set $V_{\alpha , \beta}$ of non complete rooted  (unordered)  subtrees of $G=(V,E)$, of depths less than $\beta-1$ and arity less than $\alpha$. Given two different rooted unordered labelled trees, they are said \emph{equivalent} if they share the same tree structure, and this is denoted $\sim_t$.

\begin{figure}
\begin{center}

\vspace*{-.2cm}

\includegraphics[scale=.42]{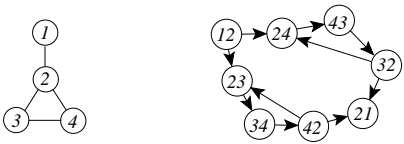}

\vspace*{-.2cm}

\caption{(left) undirected graph $G$, (right) graph $G_{1,1}$.
 }
 \label{fig:enumeration}
\end{center}
\end{figure}

On this  set $V_{\alpha , \beta}$, we define a directed graph with edge set $E_{\alpha , \beta}$ as follows: $R_0 \in V_{\alpha , \beta}$ is connected to $R_1 \in V_{\alpha , \beta}$ if ``the tree $R_1$ extends the tree $R_0$ one generation further'', i.e., if and only if (a) the first $\beta-2$ generations of $R_1$ are exactly equal to one of the  complete subtree of $R_0$ rooted at a child of the root of $R_0$, and (b) the nodes of depth $\beta-1$ of $R_1$ are distinct from the nodes in $R_0$. This defines a graph $G_{\alpha , \beta}=(V_{\alpha , \beta},E_{\alpha , \beta})$ (see \myfig{enumeration}). Similarly we define a graph $H_{\alpha,\beta} = (W_{\alpha , \beta},F_{\alpha , \beta})$ for the graph $H$.

For a $\beta$-tree-walk, the root with its $\beta-1$-descendants must have distinct vertices and thus correspond exactly to elements of $V_{\alpha,\beta}$. We let $k^\mathcal{T}_{\alpha,\beta,\gamma}(G,H,R_0,S_0)$ denote the same kernel as defined in \eq{kernel}, but restricted to tree-walks that start respectively start with $R_0$ and $S_0$. Note that if $R_0$ and $S_0$ are not equivalent, then $k^\mathcal{T}_{\alpha,\beta,\gamma}(G,H,R_0,S_0)=0$

We obtain the following recursion for all $R_0 \in V_{\alpha , \beta}$ and 
and $S_0 \in W_{\alpha , \beta}$ such that $R_0 \sim_t S_0$:
$$k^\mathcal{T}_{\alpha,\beta,\gamma}(G,H,R_0,S_0) = 
\sum_{p=0}^\alpha \lambda^p \nu^{(p-1)_+} \!\!\!\! \sum_{ 
\begin{array}{c} R_1,\dots,R_p \subset \mathcal{N}_{G_{\alpha , \beta}}(R_0) \\ R_1,\dots,R_p \mbox{ disjoint}
\end{array}
}
 \sum_{ 
\begin{array}{c} S_1,\dots,S_p \subset \mathcal{N}_{H_{\alpha , \beta}}(S_0) \\ S_1,\dots,S_p \mbox{ disjoint}
\end{array}
}  $$

\vspace*{-.25cm}

$$ \hspace*{3cm} k_\mathcal{A}(A_{r(R)},B_{r(S)}) 
\frac{ k_\mathcal{B}^{ \cup_{i=1}^p R_i | R_0, \cup_{i=1}^p S_i | S_0  }(K,L) }
{\prod_{i=1}^p k_\mathcal{B}^{R_i,S_i}(K,L) }
\left( \prod_{i=1}^p k^\mathcal{T}_{\alpha,\beta,\gamma-1}(G,H,R_i,S_i) \right) .
 $$
 Note that if any of the trees $R_i$ is not equivalent to $S_i$, it does not contribute to the sum.
 The recursion is initialized with 
 $k^\mathcal{T}_{\alpha,\beta,\gamma}(G,H,R_0,S_0) = \lambda^{|R_0|} \nu^{\ell(R_0)} k_\mathcal{A}(A_{R_0},B_{S_0})
 k_\mathcal{B}(K_{R_0},L_{S_0})$ and the final kernel is obtained as
  $ \textstyle k^\mathcal{T}_{\alpha,\beta,\gamma}(G,H) = \sum_{R_0 \sim_t S_0}  k^\mathcal{T}_{\alpha,\beta,\gamma}(G,H,R_0,S_0)$.
 
Note that we may reduce the computational load by considering a set of trees of smaller arity in the previous recursions; i.e., we consider $V_{1,\beta}$ instead of $V_{\alpha,\beta}$ with tree-kernels of arity $\alpha>1$.

 \subsection{Computational complexity}
 The complexity of computing one kernel between two graphs is linear in $\gamma$ (the order of the tree-walks), and quadratic in the size of $V_{\alpha , \beta}$ and $W_{\alpha , \beta}$. However, those sets have exponential size in $\beta$ and $\alpha$ in general. And thus, we are limited to small values (typically $\alpha\leqslant 3$ and $\beta \leqslant 6$) which are sufficient for good classification performance (see \mysec{simulations}).
For example, for the handwritten digits we use in simulations, the average number of nodes in the graphs are $18 \pm 4$ , while the average cardinal of $V_{\alpha,\beta}$ is  $37 \pm 13$ ($\alpha=1$, $\beta=4$),  and  $ 70 \pm 44$ ($\alpha=2$, $\beta=4$).

\section{Application to handwritten character recognition}

\label{sec:simulations}
We have tested our new kernels on the task of isolated handwritten character recognition, handwritten arabic numerals (MNIST dataset) and Chinese characters (ETL9B dataset). We selected the first 100 examples for the ten classes in the MNIST dataset, while for the ETL9B dataset, we selected the five hardest classes to discriminate among the 3,000 (by computing distances between class means) and then selected the first 50 examples per class. Our learning task it to classify those characters; we use a one-vs-one multiclass scheme with 2-norm support vector machines~\cite{ShaweCristi04}.

\begin{figure}
\begin{center}

\vspace*{-.46cm}

\includegraphics[scale=.3]{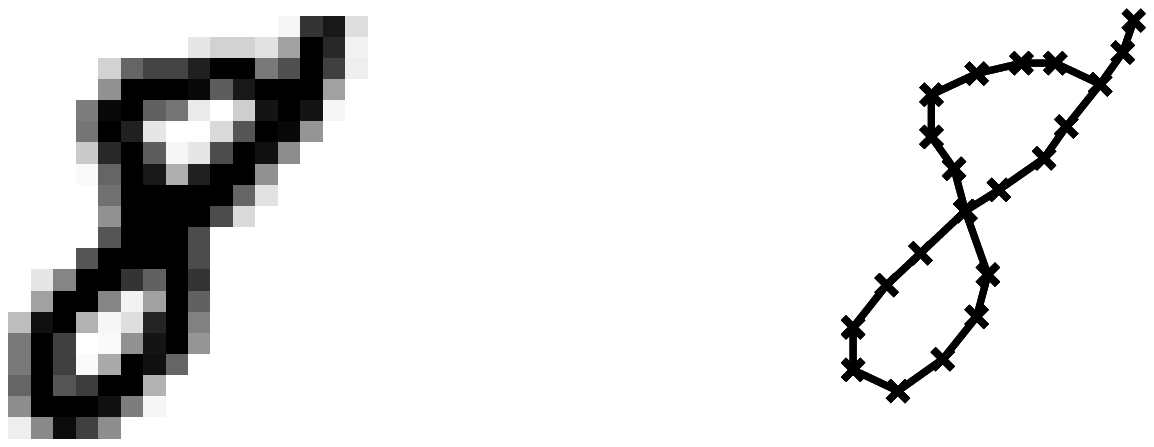} \hspace*{0cm}
\includegraphics[scale=.3]{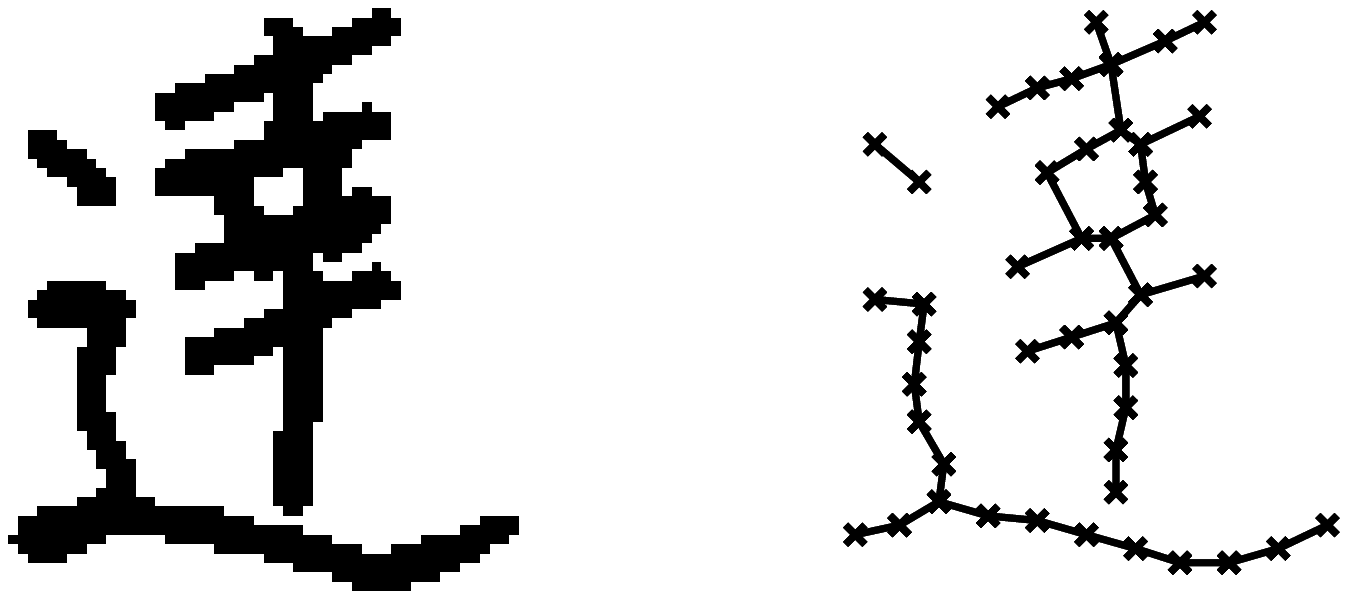}

\vspace*{-.5cm}

\caption{For digits and Chinese characters: (left) Original characters, (right) thinned and subsampled characters
 }
 \label{fig:characters}
\end{center}
\end{figure}

We consider characters as drawings in $\rb^2$, which are sets of possibly intersecting contours. Those are naturally represented as undirected planar graphs. We have thinned and subsampled uniformly each character to reduce the sizes of the graphs (see two examples in \myfig{characters}).

The kernel on positions is $k_\mathcal{X}(x,y) = \exp( - \tau \|x-y\|^2 ) + \kappa \delta(x,y)$, but could  take into account different weights on horizontal and vertical directions. We add the positions from the center of the bounding box as features, to take into account the global positions, i.e., we use $k_\mathcal{A}(x,y) = \exp( - \upsilon \|x-y\|^2)$. This is necessary because the problem of handwritten character recognition is not globally translation invariant.

In this paper we have defined a family of kernels, corresponding to different values of the following free parameters (shown with their possible values):
\begin{center}
\begin{tabular}{|l|l|l|}
\hline
$\alpha$ & arity of  tree-walks & 1, 2 \\
$\beta$ &  order of tree-walks & 1, 2, 4, 6 \\
\hline
$\gamma$ & depth of tree-walks & 1, 2, 4, 8, 16, 24 \\
$\lambda$  & penalization on number of nodes & 1  \\
$\nu$      & penalization on number of leaf nodes &  .1, .01 \\
\hline
$\tau$ & bandwidth for kernel on positions & .05, .01, .1 \\
$\kappa$ &  ridge parameter  & $.001$ \\
$\upsilon$ &  bandwidth  for kernel on attributes & .05, .01, .1 \\
\hline
\end{tabular}
\end{center}
The first two sets of parameters ($\alpha,\beta,\gamma,\lambda,\nu$) are parameters of the graph kernel, independent of the application, while the last set ($\tau,\kappa,\nu$) are parameters of the kernels for attributes and positions. Note that with only a few important scale parameters ($\tau$ and $\nu$), we are able to characterize complex interactions between the vertices and edges of the graphs. In practice, this is very important, to avoid considering many distinct parameters for all sizes and topologies of subtrees.

 In simulations, we performed two loops of 5-fold cross-validation: in the outer loop, we consider 5 different training folds with their corresponding testing folds. On each training fold, we consider all possible values of $\alpha$ and $\beta$. For all of those values, we select all other parameters (including the regularization parameters of the SVM) by 5-fold cross-validation (the inner folds). Once the best parameters are found only by looking at the training fold, we train on the whole training fold, and test on the testing fold. We output the means and standard deviations of the testing errors for each testing fold. We compare the performance with the Gaussian-RBF kernel with bandwidth learned by cross-validation in the same way in \myfig{results}.

\begin{figure}
\begin{center}
\begin{tabular}{|l||l|l|l|l||l|}
\hline
& $\beta=1$ & $\beta=2$ &  $\beta=4$ & $\beta=6$ & RBF \\
\hline
MNIST - $\alpha=1$ & $ 8.6 \pm 1.3$ & $ 7.0 \pm 2.1 $ & $\mathbf{ 5.4 \pm 1.0} $ & $ 5.6 \pm 2.2$ & $ 9.3 \pm 2.1 $\\
MNIST - $\alpha=2$ &$ 7.3 \pm 2.0 $ & $ 6.8 \pm 1.0 $ & $ 6.6 \pm 1.2$ & $5.7 \pm 1.0$ & -  \\
\hline
ETL9B - $\alpha=1$& $38.8 \pm 7.0$ & $ 34.8 \pm 7.0$ & $ 30.0 \pm 8.0 $ & $31.2 \pm 7.8$  & $48.4 \pm 5.9$ \\
ETL9B - $\alpha=2$ & $45.2 \pm 9.9$ & $29.2 \pm 10.7$ & $31.2 \pm 6.7$ & $\mathbf{25.6 \pm 4.3}$ & - \\
\hline
\end{tabular}
\end{center}

\vspace*{-.25cm}

\caption{Error rates (multiplied by 100) on handwritten character classification tasks}
\label{fig:results}
\end{figure}

These results show that our new family of kernels that use the natural structure of line drawings are outperforming the ``blind'' Gaussian-RBF kernel (error rate of $5.4$\% instead of $9.3\%$ for the MNIST digits and $25.6\%$ instead of $48.4\%$ for the ETL9B characters). Note that for arabic numerals, best performance is achieved for $\alpha=1$ (walks instead of tree-walks), which is not surprising since most digits have a linear structure (graphs are chains). On the contrary, for Chinese characters, best performance is achieved for binary tree-walks.

\section{Conclusion}
We have presented a new kernel for point clouds which is based on comparisons of local subsets of the point clouds. Those comparisons are made tractable by (a) considering subsets based on tree-walks and walks, and (b) using a specific factorized form for the local kernels between tree-walks, namely a factorization on a graphical model. 

Moreover, we have reported applications to handwritten character recognition where we showed that the kernels were able to capture the relevant information to allow good predictions from few training examples. We are currently investigating other domains of applications of points clouds, such as shape mining in computer vision~\cite{shape,suard}, and prediction of protein functions and interactions from their three-dimensional structures~\cite{protein}.

 \small
\bibliographystyle{unsrt}
\bibliography{pc}

\begin{thebibliography}{10}

\bibitem{vert}
J.-P. Vert, H.~Saigo, and T.~Akutsu.
\newblock Local alignment kernels for biological sequences.
\newblock In {\em Kernel Methods in Computational Biology}. MIT Press, 2004.

\bibitem{vert_speech}
M.~Cuturi, J.-P. Vert, O.~Birkenes, and T.~Matsui.
\newblock A kernel for time series based on global alignments.
\newblock In {\em Proc. ICASSP 2007}, 2007.

\bibitem{lodhi}
H.~Lodhi, C.~Saunders, J.~Shawe-Taylor, N.~Cristianini, and C.~Watkins.
\newblock Text classification using string kernels.
\newblock {\em Journal of Machine Learning Research}, 2:419--444, 2002.

\bibitem{harchaoui}
Z.~Harchaoui and F.~Bach.
\newblock Image classification with segmentation graph kernels.
\newblock In {\em Proc. CVPR}, 2007.

\bibitem{lecun-98}
Y.~LeCun, L.~Bottou, Y.~Bengio, and P.~Haffner.
\newblock Gradient-based learning applied to document recognition.
\newblock {\em Proc. IEEE}, 86(11):2278--2324, 1998.

\bibitem{chinese}
S.~N. Srihari, X.~Yang, and G.~R. Ball.
\newblock Offline {C}hinese handwriting recognition: A survey.
\newblock {\em Frontiers of Computer Science in China}, 2007.

\bibitem{Sun_05a}
Q.~Sun and G.~DeJong.
\newblock Feature kernel functions: Improving svms using high-level knowledge.
\newblock In {\em Proc. CVPR}, 2005.

\bibitem{shape}
S.~Belongie, J.~Malik, and J.~Puzicha.
\newblock Shape matching and object recognition using shape contexts.
\newblock {\em IEEE Trans. PAMI}, 24(24):509--522, 2002.

\bibitem{ponce}
D.~A. Forsyth and J.~Ponce.
\newblock {\em Computer Vision: A Modern Approach}.
\newblock Prentice Hall, 2003.

\bibitem{ShaweCristi04}
J.~Shawe-Taylor and N.~Cristianini.
\newblock {\em Kernel Methods for Pattern Analysis}.
\newblock Cambridge Univ. Press, 2004.

\bibitem{RamonGaert03}
J.~Ramon and T.~G\"{a}rtner.
\newblock Expressivity versus efficiency of graph kernels.
\newblock In {\em First International Workshop on Mining Graphs, Trees and
  Sequences}, 2003.

\bibitem{KashiTsudaInoku04}
H.~Kashima, K.~Tsuda, and A.~Inokuchi.
\newblock Kernels for graphs.
\newblock In {\em Kernel Methods in Computational Biology}. MIT Press, 2004.

\bibitem{mahe-tech}
P.~Mah\'e and J.-P. Vert.
\newblock Graph kernels based on tree patterns for molecules.
\newblock Technical Report CCSD-00095488, HAL, 2006.

\bibitem{kondor}
R.~I. Kondor and T.~Jebara.
\newblock A kernel between sets of vectors.
\newblock In {\em Proc. ICML}, 2003.

\bibitem{lauritzen}
S.~Lauritzen.
\newblock {\em Graphical Models}.
\newblock Oxford University Press, 1996.

\bibitem{jordanreview}
M.~I. Jordan.
\newblock Graphical models.
\newblock {\em Statistical Science}, 19:140--155, 2004.

\bibitem{caelli}
T.S. Caetano, T.M. Caelli, D.~Schuurmans, and D.A.C. Barone.
\newblock Graphical models and point pattern matching.
\newblock {\em IEEE Trans. PAMI}, 28(10):1646--1663, 2006.

\bibitem{suard}
F.~Suard, A.~Rakotomamonjy, and A.~Benrshrair.
\newblock Kernel on bag of paths for measuring similarity of shapes.
\newblock In {\em Proc. ESANN}, 2007.

\bibitem{protein}
J.~Qiu, M.~Hue, A.~Ben-Hur, J.-P. Vert, and W.~S. Noble.
\newblock A structural alignment kernel for protein structures.
\newblock {\em Bioinformatics}, 23:1090--1098, 2007.

\end{thebibliography}

\end{document}